\documentclass[11pt,letterpaper]{article}
\usepackage{ijcnlp2017}
\usepackage{times}
\usepackage{latexsym}
\usepackage{hyperref}

\usepackage{amsmath,amssymb}
\usepackage{algpseudocode,algorithm,algorithmicx}
\usepackage{graphicx,color}
\usepackage{booktabs}
\usepackage{subfig,color}
\usepackage{bm}
\usepackage{epstopdf}
\usepackage{xspace}
\usepackage[textsize=small]{todonotes}

\newcommand{\forelm}{foreLM\xspace}
\newcommand{\backlm}{backLM\xspace}

\newcommand\fnurl[2]{%
\href{#2}{#1}\footnote{\url{#2}}%
}

\ijcnlpfinalcopy

\def\ijcnlppaperid{***}

\title{Named Entity Recognition with stack residual LSTM and trainable bias decoding}

 \author{Quan Tran, Andrew MacKinlay \and Antonio Jimeno Yepes \\
         IBM Research Australia \\ {\tt quanthdhcn@gmail.com, admackin@au1.ibm.com, ayepes@au1.ibm.com} }

\date{}

\begin{document}

\maketitle

\begin{abstract}
Recurrent Neural Network models are the state-of-the-art for Named Entity Recognition (NER).
We present two innovations to improve the performance of these models.
The first innovation is the introduction of residual connections between the Stacked Recurrent Neural Network model to address the degradation problem of deep neural networks.
The second innovation is a bias decoding mechanism that allows the trained system to adapt to non-differentiable and externally computed objectives, such as the entity-based F-measure.
%to address the limitations of traditional loss functions that optimize for accuracy.
Our work improves the state-of-the-art results for both Spanish and English languages on the standard train/development/test split of the CoNLL 2003 Shared Task NER dataset.
\end{abstract}

\section{Introduction}

In Natural Language Processing, the term ``Named Entity'' refers to special information 
units such as people, organizations, location names, numerical expression~\cite{nadeau2007survey}.
Identifying the references to these special entities in text is a crucial step toward
Language Understanding. Thus, there have been many works on these areas.

Some of the early systems employed hand-crafted rules~\cite{rau1991extracting,sekine2004definition},
however, the vast majority of current systems rely on machine learning models~\cite{nadeau2007survey} such as
Conditional Random Field (CRF)~\cite{mccallum2003early}, Hidden Markov Model (HMM)~\cite{bikel1997nymble}
and Support Vector Machine (SVM)~\cite{asahara2003japanese}.
Although the traditional machine learning models do not rely on manual rules, they require a manual feature engineering process, 
which is rather expensive and dependent on the domain and language.

In recent years, Recurrent Neural Network (RNN) models such as Long-Short-Term-Memory (LSTM)~\cite{hochreiter1997long}
and Gated Recurrent Unit (GRU)~\cite{chung2014empirical} have been very successful in sequence modelings tasks,
for example, Language Modeling~\cite{mikolov2010recurrent, sundermeyer2012lstm}, 
Machine Translation~\cite{bahdanau2014neural} and Dialog Act Classification~\cite{kalchbrenner2013recurrent,tran2017hierarchical}.
One of the strengths of the RNN models is their ability to learn from basic components of text (i.e. words and characters).
This generalization capability facilitates the construction of Language Independent NER models~\cite{ma2016end,lample2016neural} 
that rely on unsupervised feature learning and a small annotated corpus.

%We based our system upon the proven LSTM-CRF architecture in these two works, and improve upon them using our novel model innovations.

One simple way of adding representational power to a neural network is layer stacking. A traditional feed forward neural network usually has three fully connected layers: an input layer, a hidden layer, and an output layer.
For a Convolutional Neural Network (CNN) or Recurrent Neural Network, the number of stacked layers might be much larger~\cite{amodei2016deep}.
One problem with this stacking scenario is the degraded representation problem~\cite{he2016deep}.
The proposed solution for this problem is the residual-identity connection~\cite{he2016deep}.
With the information from the lower-level inputs, the upper neural network layers can learn to compensate for the representation errors of lower layers. We adopt this idea for Stacking RNN, however, with a different implementation.

Most of the RNN-based models for NER and machine translation are trained with some form of maximum likelihood log-loss.
However, it is often desirable to optimize task-specific metrics\mbox{~\cite{xu2016expected}}, for example, F-measure in NER, but optimizing the F-measure directly is not trivial~\cite{busa2015online}, especially in the case of complex Deep Neural Network models.
It is even more difficult considering the way the F-measure is calculated in Named Entity Recognition in the \fnurl{CoNLL-2003 shared task}{https://www.aclweb.org/aclwiki/index.php?title=CONLL-2003_(State_of_the_art)}
, where it depends on the actual/predicted entities and~\textit{not} on each token-prediction for which the system is trained for.
Inspired by the idea of trainable decoding recently proposed in machine translation~\cite{gu2017trainable}, we introduce a trainable \textit{percentage bias decoding} system that manipulates the outputs of a base system trained with normal log-loss to adapt to a new objective.
Our trainable bias decoding system also bears similarity to the thresholding technique~\cite{lipton2014optimal}, traditionally used to maximize F-measures given a classifier.
The proposed decoding system is trained directly on the externally computed F-measures (which relies on the the CoNLL evaluation script) using finite different gradient.

In the next sections, we describe the proposed innovations with detailed motivations and discussions.
Results show that our proposed innovations improve the NER state-of-the-art for the English and Spanish languages in the CoNLL-2003 shared task data set.

\section{Models}
%!TEX root = ijcnlp2017.tex
\begin{figure*}
\centering
\includegraphics[scale=0.5]{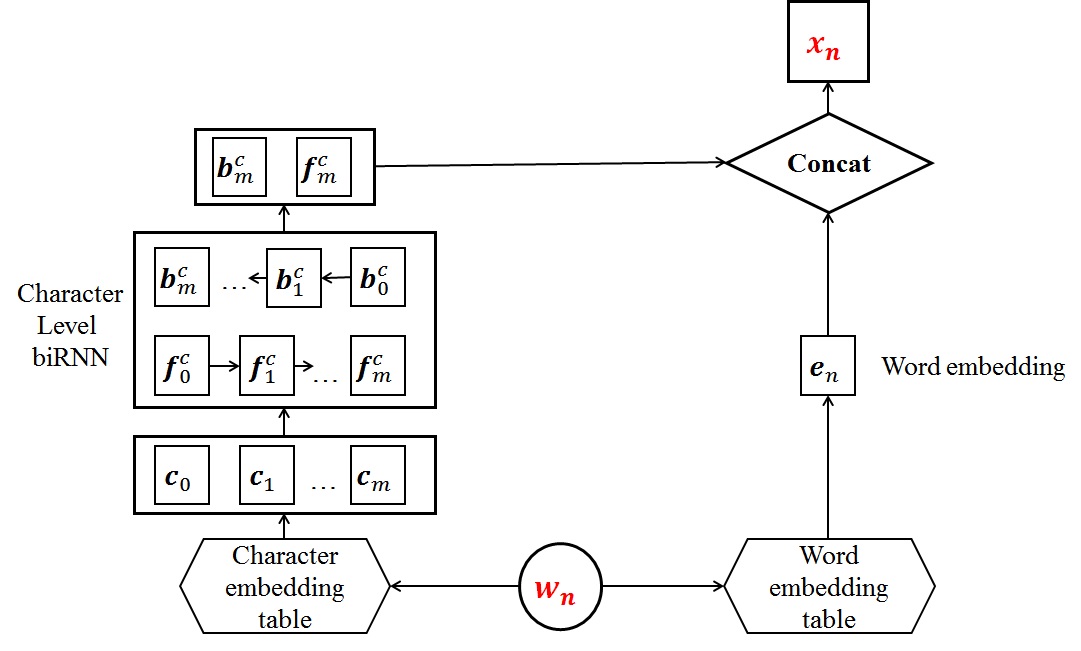}
\caption{Extracting word features with word embeddings and character level biRNN}
\label{fig:char}
\end{figure*}

We describe first our RNN-CRF base architecture and then we describe our two modelling innovations: the Stack Residual RNN and the bias decoding.

\subsection{The base RNN-CRF architecture}

Our system is built upon the RNN-CRF architecture for Named Entity Recognition. 
Let us denote the input sequence of words as $w_0,...,w_n$. 
In general, the RNN component encodes the words into a sequence of hidden vectors $h_0,...,h_n$. 
This sequence of hidden vectors is then treated as features for a linear-chain CRF layer. 
The training objective will then be the log-likelihood of the correct sequence.

Following \newcite{lample2016neural,ma2016end,yang2016multi}, we employ the character level information and word-embeddings as features in NER. 
Similar to \newcite{lample2016neural}, in our system, character information is encoded using a bi-directional RNN (biRNN) over characters.
Given a word $w_k \in w_0,...,w_n$ with $m$ characters, let us denote the character-embedding sequence of this word as $\textbf{c}_0,...,\textbf{c}_m$, the biRNN function as $\rho$ and the concat function as $\psi$. 
The character embedding representation $\textbf{h}^c_k$ of word $w_k$ is calculated using a biRNN as in Equation \ref{eq:char}, in which $\textbf{f}_0,...,\textbf{f}_m$ and $\textbf{b}_0,...,\textbf{b}_m$ denote the hidden units in the forward and backward RNNs respectively.

\begin{align}
    \label{eq:char}
    \begin{split}
        \textbf{f}_0,...,\textbf{f}_m &= \rho(\textbf{c}_0,...,\textbf{c}_m) \\
        \textbf{b}_0,...,\textbf{b}_m &= \rho(\textbf{c}_m,...,\textbf{c}_0) \\
        \textbf{h}^c_k &= \psi([\textbf{f}_m,\textbf{b}_m]) 
    \end{split}
\end{align}

The feature vector $\textbf{x}_k$ of word $w_k$ is then the concatenation of $\textbf{h}^c_k$ and the traditional word-embedding $\textbf{e}_k$ as shown in Equation \ref{eq:feat}.
Figure \ref{fig:char} shows the feature extraction procedures. 
All the parameters of the biRNN as well as the embedding tables are jointly trained with other component of the model. 
The word embedding table is initialized with a pre-trained embedding table.

\begin{equation}
    \label{eq:feat}
    \textbf{x}_k = \psi([\textbf{h}^c_k,\textbf{e}_k])
\end{equation}

The most simple architecture would be a one-directional RNN over the word features: $h_0,...,h_n = RNN(x_0,...,x_n)$. However, it has been shown to be beneficial to have a bidirectional RNN over the input layer, as a bidirectional RNN captures both the left and the right context of a word: $h_0,...,h_n = \rho(x_0,...,x_n)$.

The final sequence of hidden vectors $\textbf{h}_0,...,\textbf{h}_n$ is treated as the features for a linear-chained CRF layer. 
Similar to \newcite{lample2016neural}, the observation scores $\lambda$ are calculated with a linear transformation from the hidden vectors, as show in Equation \ref{eq:featmap}, where $\lambda_i$ is the vector observation scores for all the labels in $i$th time-step, $\textbf{W}_p$ is an $l \times d$ weight matrix, and $\textbf{b}_p$ is a bias vector of size $l$ (with $d$ is the size of vector $\textbf{h}_i$), and $l$ is the size of the label set $\mathbb{Y}$ (including the special sequence begin and end labels).

%% AJY: Quan, indices i and p are not defined. W should be defined as well
%% Quan: changed 

\begin{align}
	\label{eq:featmap}
	\lambda_i = \textbf{W}_p \textbf{h}_i + \textbf{b}_p
\end{align}

Given a sequence of input words $S=w_0,...,w_n$, the score of a particular sequence of labels $Y=Y[0],...,Y[n]$ is calculated using the observation scores $\lambda$ and the transition scores $\delta$ as in Equation \ref{eq:CRF}, where $\delta$ is a square matrix of dimension $l \times l$, $\delta(Y[j],Y[j+1])$ denotes the transition score between the label in position $j$ and the label in position $j+1$ in sequence $Y$, and $\lambda_i(Y[i])$ is the observation score of $i$-th label $Y[i]$.

\begin{align}
\label{eq:CRF}
\begin{split}
	\zeta(Y,S) &= \sum_{i:[0..n]}\lambda_i(Y[i]) \\
		    &+\sum_{j:[0..n-1]}\delta(Y[j],Y[j+1])
\end{split}
\end{align}

The probability of a sequence $Y$ is calculated using a softmax over all the possible sequences $\mathbb{Y}$ (Equation \ref{eq:soft}).

\begin{align}
	\label{eq:soft}
	Pr(Y|S) = \frac{e^{\zeta(Y,S)}}{\sum_{\bar{Y}\in \mathbb{Y}} e^{\zeta(\bar{Y},S)}}
\end{align}

During training, we maximize the log-likelihood of the correct sequence $Y_c$.
The loss function $\mathbb{L}$ is defined in Equation \ref{eq:loss}.
Because we employ a linear-chain CRF, the term $log(\sum_{Y'\in \mathbb{Y}} e^\zeta(Y',S))$ in Equation \ref{eq:loss} can be efficiently calculated with dynamic programming.

\begin{align}
    \label{eq:loss}
    \mathbb{L}(Y_c) = \zeta(Y_c,S) - log(\sum_{\bar{Y}\in \mathbb{Y}} e^\zeta(\bar{Y},S))
\end{align}

%% AJY: Quan, check inline math expressions
During decoding, the best sequence can be found using the Viterbi algorithm.
The original Viterbi decoding algorithm builds an $l \times n$ score table $\xi$ 
in which $l$ is the size of the label set (including the beginning and end labels) and $n$ is the length of the sequence.
$\xi_j(y_i)$ denotes the score of the most probable partial path (up to position $j$) with position $j$ having the label $y_i$. 
$\xi_j(y_i)$ is calculated using dynamic programming as in Equation \ref{eq:dynamic}. 

\begin{align}
    \label{eq:dynamic}
    \xi_j(y_i) = \sum_{y_k \in \mathbb{Y}}(\xi_{j-1}(y_k) +\delta(y_k,y_i)) + \lambda_j(y_i)
\end{align}

\begin{figure*}
\centering
\includegraphics[scale=0.5]{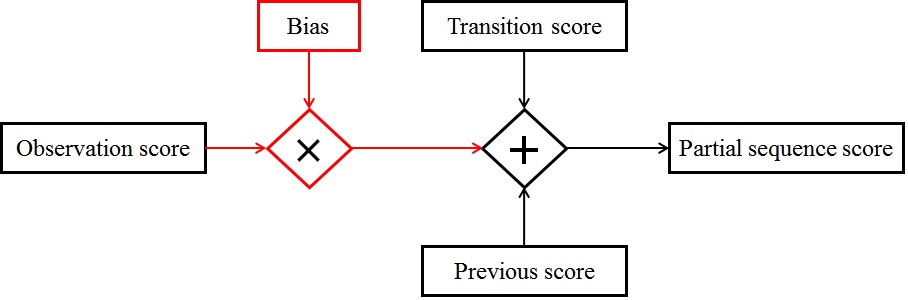}
\caption{The application of percentage bias to Viterbi decoding}
\label{fig:bias}
\end{figure*}

\begin{figure}
\centering
\includegraphics[scale=0.5]{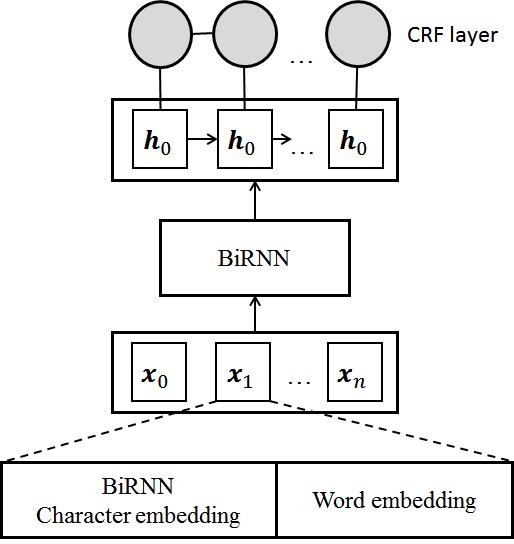}
\caption{The RNN-CRF architecture}
\label{fig:CRFRNN}
\end{figure}

%% AJY: Quan, check $\hat{Y}$
%% Quan: checked and changed 
At the end of the decoding process, sequence $\hat{Y}$ is predicted by selecting the best score at the end of the sequence $j=n$ and then completing the sequence with a backward pointer (Equation \ref{eq:viterbi}). 
Figure \ref{fig:CRFRNN} depicts the whole RNN-CRF architecture

\begin{align}
    \label{eq:viterbi}
    \begin{split}
        \hat{Y}[n]  &= \arg\max_{y_{i}} \xi_n(y_i) \\
        %\hat{Y}[0],...,\hat{Y}[n-1] &= Backward(\hat{Y}[n])
		\hat{Y}[n-1] &= \arg\max_{y_{k}}(\xi_{j-1}(y_k) +\delta(y_k,\hat{Y}[n]))
    \end{split}
\end{align}

\subsection{Stacked Residual RNN}

\begin{figure*}
\centering
\includegraphics[scale=0.55]{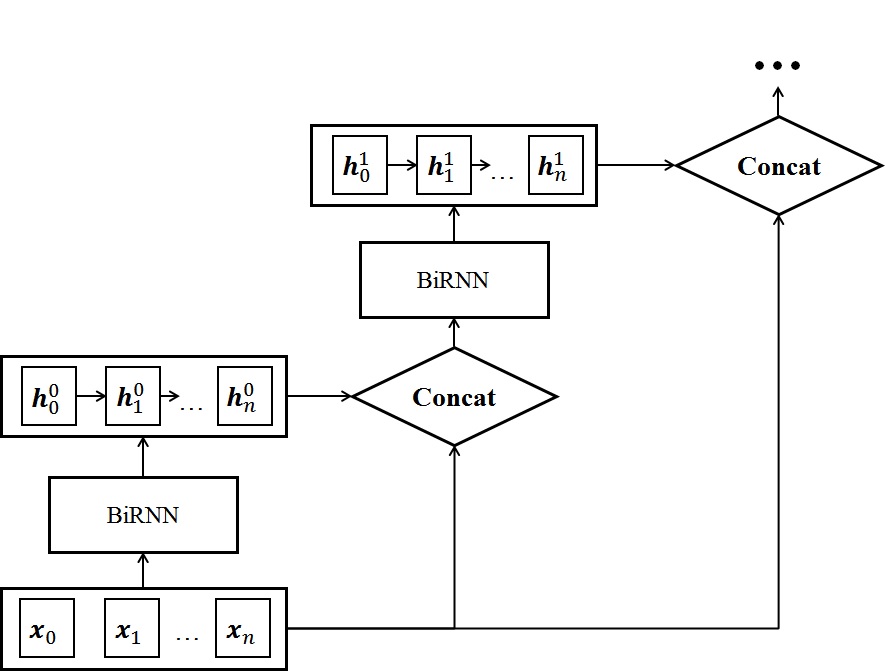}
\caption{Feature learner mixture with residual connection}
\label{fig:stack}
\end{figure*}

A traditional way of adding more representational power to a neural network is layer stacking.
RNN stacking has been successfully used in a lot of works~\cite{amodei2016deep}.
However, stacking layers of neural networks suffers from the \textit{degradation problem}~\cite{he2016deep}.
This is due to the difficulty in training a lot of stacked layers and fit these layers to desired underlying mappings, which leads to representational degradation.

The solution proposed to this problem, the \textit{residual connection} {\cite{he2016deep, prakash2016neural} tries to create shortcuts between non-consecutive layers.
However, the original additional residual connection (adding the input vector to the hidden representation) adds several constraints on the dimensionality of the hidden and input layers, which might require vector clipping~\cite{prakash2016neural}, and it might lead to a loss of information.

In the original residual connection proposed for image recognition, the residual information is summed to the output of the upper layers ($\mathbb{F}(x)+ x$). In our proposal, we want the upper layer of a neural network to have direct access to the original input, thus, the original input is now appended to the output of the lower layers instead of being summed.
With this formulation, there is no dimensionality restriction, and furthermore, we argue that our proposed residual connection can be used to mix feature learners of different complexity (Figure \ref{fig:stack}).
For example, when equipped with our proposed residual connection, the top neural network layer can act like a shallow one-layer feature learner.
The two top layers can act like a deeper two-layer feature learner.
Equation set \ref{eq:stack} shows the exact formulation of our proposed residual connection within the Stack RNN. Similar to the Equation \ref{eq:char}, we denote the biRNN function as $\rho$ and the concat function as $\psi$.
This modelling procedure is depicted in Figure \ref{fig:stack}.

\begin{align}
    \label{eq:stack}
    \begin{split}
        \textbf{h}^0_0,...,\textbf{h}^0_n &= \rho(\textbf{x}_0,...,\textbf{x}_n) \\
        \hat{\textbf{h}}^0_0,...,\hat{\textbf{h}}^0_n &= \psi([\textbf{x}_0,\textbf{h}^0_0]),...,\psi([\textbf{h}^0_n, \textbf{x}_n]) \\
        \textbf{h}^1_0,...,\textbf{h}^1_n &= \rho(\hat{\textbf{h}}^0_0,...,\hat{\textbf{h}}^0_n) \\
        \textbf{h}^M_0,...,\textbf{h}^M_n &= \rho(\hat{\textbf{h}}^{M-1}_0,...,\hat{\textbf{h}}^{M-1}_n)
    \end{split}
\end{align}
%\amcomm{what are $\psi$ and $\rho$?}
%Quan: fixed, redefined both.

\subsection{The bias decoding}

%Traditionally, deep neural networks in NER are trained with some forms of log-likelihood.
%which typically optimizes the accuracy of the trained system.
%However, the evaluation process is typically performed using a different measure, e.g. for the CoNLL 2013 Shared Task NER dataset, the evaluation is
%performed by an external script using entity-based F-measure. Training directly on the entity-based F-measure is not trivial as it is non-differentiable and %externally computed evaluation measures. It is usually beneficial to train the neural network on the task-specific evaluation metric
%\mbox{~\cite{xu2016expected}}.
%Thus, we look into a way of train our model to adapt to the externally calculated 

Usually NER systems are evaluated with some form of F-measure. 
For example, for the CoNLL 2013 Shared Task NER dataset, the evaluation is performed by an external script using entity-based F1-measure. 
Although it has been noted that training on the evaluation metric is beneficial \mbox{~\cite{xu2016expected}}, most of the deep models for NER are trained with log-likelihood. The main reason for this discrepancy is the difficulty in training with F-measures. 
Instead of trying to train on F-measure directly, we look into a hybrid solution where we train a model on log-likelihood first, and then use a simpler ``adaptation model'' to manipulate the output of the base model to fit it to the F-measure.

Inspired by Machine Translation research on decoding with trainable noise~\cite{gu2017trainable}, we explore the possibility of adding trainable noise to the Viterbi decoding process.
Analogously to the traditional threshold technique for maximizing the F1 score in binary classification, we introduce a simple \textit{percentage noise} to the decoding process.
That is, during the construction of the score table $\xi$ (Equation \ref{eq:dynamic}), a label-specific percentage bias is added to the calculation as in Equation \ref{eq:bias}.
Figure \ref{fig:bias} shows the application of this bias to the Viterbi decoding.

\begin{align}
    \label{eq:bias}
    \xi_j(y_i) = \sum_{y_k \in \mathbb{Y}}(\xi_{j-1}(y_k) +\delta(y_k,y_i)) + b_y \lambda_j(y_i) 
\end{align}

To test this new percentage bias idea, we perform a quick experiment, where we limit the use of bias to the most numerous class in the CoNLL tag set, class O (words that do not belong to any entity).
We search for the best bias $b_O$ from the range of $[0.5,1.5]$ using a value loop with step of $0.1$.
For each value of $b_O$, we calculate the F1-measure on the validation set, and choose the value with the highest F1.
We use our trained model based on the Stack Residual architecture above as the base probabilistic model.
We find that the best $b_O$ value is $1.1$ (a value of $1.0$ means without any bias). Using this $b_O$ bias for the test data yields the F1-measures of $91.22$ compared to the original score of $91.07$ in the test set.
This experiment supports our claim that the base model trained with log-likelihood might not optimize well on a different performance measure, and adding this percentage bias noise is really beneficial.

%Motivated by this result, we explored ways to properly train the bias noise.
We extended this idea treating the biases as parameters.
Thus the trainable bias decoding system has the number of parameters equal to the number of classes.
Training with gradient descent with CoNLL's entity-based F1 loss is rather difficult, as it is hard to calculate the exact gradient.
This is solved using the numerical gradient methods as an approximation, which is shown in Equation \ref{eq:grad}.

\begin{align}
    \label{eq:grad}
    f'_b \sim \frac{f(b+\epsilon) - f(b-\epsilon)}{2\epsilon} 
\end{align}

The training procedure is then very similar to stochastic gradient descent.
Details on the choice of hyper parameters and other experimental settings are presented in the Experiment section.

\section{Experiments}
%!TEX root = ijcnlp2017.tex
\subsection{Dataset and Experimental settings}

We have prepared and evaluated the proposed methods on the English and Spanish sets of the CoNLL 2003 NER data set\footnote{http://www.cnts.ua.ac.be/conll2003/ner}\mbox{~\cite{tjong2003introduction}}.
We have reused the training, development and test set configuration of the CoNLL-2003 Shared Task in our study.

The training set has been used to train the system using several hyperparameter configurations, the development set has been used to select the best configuration and the reported performance of the final system is based on the test set. The Spanish dataset has 8323/1915/1517 sentences in train/dev/test sets respectively. The English dataset is almost twice as large with 14041/3250/3453 sentences in train/dev/test set. For all of our models, the word-embedding size is set to 100 for English and 64 for Spanish. The hidden vector size is 100 for both English and Spanish sets without the LM embeddings.
With the LM embeddings, the hidden vector size is changed to 300 for English.
We trained the model with Stochastic Gradient Descent (SGD) with momentum, using the learning rate of 0.005. For the bias decoding, the $\epsilon$ hyperparameter for each update is randomly chosen from a range of $[0.01,...,0.1]$ with step-size of 0.01. 
Because the base model trained with sequence level log-likelihood fits very well on the training set, the gradient calculated with Equation \ref{eq:grad} might be every small, thus we opt to calculate the finite difference with respect to the loss: $log2(1-F1/100)$ instead of the $F1$ to boost the gradient information in the points where F1 is very close to 100 (perfect classification).
The learning rate for bias training is also set to 0.005. 
Statistical significance has been determined using a randomization version of the paired sample t-test~\citep{cohen1996empirical}.

%For each language, the results are compared to state of the art reported results and to a baseline system consisting of the method provided by\cite{lample2016neural} using the word embeddings mentioned earlier.

We first conduct several series of experiments to confirm the effectiveness of our two proposed ideas: the Stack Residual RNN and the bias decoding, and the new Language Model embedding in sub-section~\ref{sec:comp}. The second sub-section:~\ref{sec:comp2} compares our method with state-of-the-art results.

\subsection{Component Analysis}
\label{sec:comp}

\textbf{Adding stack Residual RNN} \\
Due to computational complexity, there is a practical limit on how many RNN layers we can stack. In this series of experiment, we tested our model without Stacked Residual RNN, and with 2, 3 and 4 Stacked layers. 
The word embeddings are initialized using the pre-trained word vectors described below.
The result of this series of experiements is presented in Table~\ref{tab:ana_stack}.

\begin{table*}
\centering
\begin{tabular}{p{6cm} p{3cm} p{3cm}}
\hline \bf System & \bf F1 English & \bf F1 Spanish \\ 
\hline
CRF-RNN no Stack Residual & 90.43 & 85.41\\
\hline
CRF-RNN 2 Stack Residual & 90.72 & 85.88\\
CRF-RNN 3 Stack Residual & \textbf{91.07} $\star$& \textbf{86.24} $\star$\\
CRF-RNN 4 Stack Residual & 91.02 $\star$ & 85.51\\
\hline
\end{tabular}
\caption{Analysis of the Stack Residual Component. $\star$ indicates significance ($p<0.05$) versus CRF-RNN no Stack Residual.}
\label{tab:ana_stack}
\end{table*}

From the result, we can see that the performance seems to increase as we add more stacked layers, and peak at three before dropping. We continue to analyze other components using 3 Stacked Residual Layers of CRF-RNN as the base model, we call this model \textit{3 Res-RNN} for short.

For English, the 3 and 4 stacked layer improvements are significant ($p < 0.025$) compared to the baseline model and between the stacked layer models, the improvement between 2 and 3 layers is significant ($p < 0.035$).

For Spanish, the 3 stacked layer improvement is significant ($p < 0.03$), with respect to the baseline model. Improvement between the 3 stacked layer and the 4 stacked layer models is significant ($p < 0.03$).

\textbf{Adding Language Model Embedding} \\
Pre-trained word embeddings have shown useful in Natural Language Processing tasks, but provide information about the word but not about its context.
Previous work has explored using language models in addition to word embeddings~\cite{peters2017semi} with positive results.
We have evaluated our system using pre-trained language models using the 3 Stacked Residual Layer configuration.
First, we test the models with forward-only LM embeddings (\forelm), then we test the model with both forward and \backlm (\backlm).
The result of this series of experiments is presented in Table~\ref{tab:ana_LM}.

\begin{table*}
\centering
\begin{tabular}{p{6cm} p{3cm} p{3cm}}
\hline \bf System             & \bf F1 English & \bf F1 Spanish \\ \hline
3 Res-RNN                     & 91.07          $\star$& \textbf{86.24} $\star$\\
\hline
3 Res-RNN+\forelm          & 91.43          $\star$& 86.13 $\star$\\
3 Res-RNN+\forelm+\backlm & \textbf{91.66} $\star$& 85.83\\
\hline
\end{tabular}
\caption{Analysis of the Language Model Embedding. $\star$ indicates significance ($p<0.05$) versus CRF-RNN no Stack Residual in Table~\ref{tab:ana_stack}.} 
\label{tab:ana_LM}
\end{table*}

The gain from the LM embedding is not consistent. It seems to work very well with English, where it improves performance substantially even though this improvement is not specially significant. However, the LM does not improve the performance at all in Spanish. Adding the \forelm and \backlm significantly decreases performance.

\textbf{Adding Bias Decoding} \\
We test the bias decoding on models with and without LM embeddings, with results shown in Table~\ref{tab:ana_bias}.
The bias-decoding increases the performance across the board, however the performance increases are not consistent. The increases are notable on some cases (3 Res-RNN + bias on both English and Spanish, 3 Res-RNN + \forelm + \backlm + bias for Spanish), while in some cases the increases are minimal (3 Res-RNN + \forelm + bias on both English and Spanish, 3 Res-RNN + \forelm + \backlm + bias on English).

\begin{table*}
\centering
\begin{tabular}{lrr}
\hline \bf System & \bf F1 on English & \bf F1 on Spanish \\ \hline
3 Res-RNN                                   & 91.07 $\star$& 86.24$\star$\\
3 Res-RNN + \forelm                      & 91.43 $\star$& 86.13$\star$\\
3 Res-RNN + \forelm + \backlm        & 91.66 $\star$& 85.83\\
\hline
3 Res-RNN + bias                            & 91.23          $\star$$\dagger$& \textbf{86.31} $\star$\\
3 Res-RNN + \forelm + bias               & 91.45          $\star$& 86.14 $\star$\\
3 Res-RNN + \forelm + \backlm + bias & \textbf{91.69} $\star$& 86.00 $\star$$\dagger$\\
\hline
\end{tabular}
\caption{Analysis of the bias decoding. $\star$ indicates significance ($p<0.05$) versus CRF-RNN no Stack Residual in Table~\ref{tab:ana_stack}. $\dagger$ indicates significance versus the configuration with no bias.}
\label{tab:ana_bias}
\end{table*}

For English, adding bias to the 3 Res-RNN without LM yields a significant improvement ($p < 0.013$), while for Spanish, the boost from adding bias to the 3 Res-RNN + \forelm + \backlm model is significant ($p < 0.011$).

\subsection{External Knowledge Learning}

\subsubsection{Word embedding}

%%% What did we use for English?
English word embedding was obtained from Word2vec-api\footnote{https://github.com/3Top/word2vec-api/blob/master/README.md}. The embedding dimension is 100 and it was trained using GloVe with AdaGrad.
For the generation of Spanish word embeddings we followed~\newcite{lample2016neural}, using
Spanish Gigaword Third Edition\footnote{https://catalog.ldc.upenn.edu/ldc2011t12} as corpus with an embedding dimension of 64, a minimum word frequency cutoff of 4 and a window size of 8.

\subsubsection{Language Modeling}

In some experiments, we used both forward and backward language models.
The English forward language model was obtained from TensorFlow\footnote{https://github.com/tensorflow/models/tree/master/lm\_1b} using the One Billion Word Benchmark\footnote{https://github.com/ciprian-chelba/1-billion-word-language-modeling-benchmark}~\cite{chelba2013one} and has a perplexity of 30.
As the code generating this pre-trained model is not available, we made use of a substitute which produces a higher perplexity language model.
For the backward English language model and the Spanish forward and backward ones, they were generated using an LSTM based baseline\footnote{https://github.com/rafaljozefowicz/lm}~\cite{jozefowicz2016exploring}.
This code estimates a forward language model and was adapted to estimate a backward language model.
Language models were estimated using the One Billion Word benchmark.
The vocabulary for the backward English model is the same as the pre-generated forward model.
The perplexity for estimated backward English language model is 46; despite the discrepancy in perplexity with the forward language model the performance using this language model still improves the named entity recognition task.
The vocabulary for the Spanish language models was generated using tokens with frequency $>$ 2. The perplexity for the forward and backward Spanish language models are 56 and 57 respectively.

\subsection{Comparative performance}
\label{sec:comp2}

\begin{table*}
\centering
\begin{tabular}{lrr}
\hline \bf System & \bf F1 on English & \bf F1 on Spanish \\
\hline
CRF-RNN no Stack Residual & 90.43 & 85.41\\
\hline
 \cite{passos2014lexicon} & 90.05 & -- \\
 \cite{santos2015boosting} & -- & 82.21 \\
 \cite{gillick2015multi} & 84.57 & 81.83 \\
 \cite{lample2016neural} & 90.94 & 85.75 \\
 \cite{ma2016end} & 91.21 & -- \\
\hline
3 Res-RNN + bias & 91.23 & \textbf{86.31}\\
3 Res-RNN + \forelm + bias & 91.45 & 86.14\\
3 Res-RNN + \forelm + \backlm + bias & \textbf{91.69} & 86.00\\
\hline
\end{tabular}
\caption{Compare our model with systems with comparable experimental settings}
\label{tab:ana_res}
\end{table*}

%\amcomm{We should have significance figures in these results tables – I think you have the code for this and they're good?}
Table \ref{tab:ana_res} shows the performance of our best systems compared to the state-of-the-art results on ConLL dataset. We focus our comparison to the systems with the same experimental setups (standard train/val/test split, without the use of external label data). The best previous systems~\cite{ma2016end, lample2016neural} are based upon a similar architecture (CRF-RNN) to ours. \newcite{lample2016neural} employed LSTM for character-based embedding, while \newcite{ma2016end} employed CNN for character-based embedding\footnote{There are several other works reporting very strong result on English NER: Chiu et al.~(91.62)~\shortcite{chiu2015named}, Yang et al.~(91.20)~\shortcite{yang2016multi} and Peter et al.(91.93)~\shortcite{peters2017semi}, however, these results are not comparable to ours due to the difference in experimental setup~\cite{ma2016end}.}. Overall, we achieve state-of-the-art results on both English and Spanish.

%% Mention the differences in performance of the different approaches

\section{Discussion}
%!TEX root = ijcnlp2017.tex

%% Look at what is possible to explore from the proposed approaches.

%% 1. Performance of the proposed methods compared to state of the art methods

%% 2. Why LM does not work for the Spanish set.

Overall, our model achieves the state-of-the-arts for both English and Spanish Named Entity Recognition.
For Spanish, our base model with three layers of Stacked Residual RNN already outperforms the current state-of-the-art. 

From the results above, we can see that our innovations, the Stacked Residual connection and bias decoding
consistently improve the performance across both data sets.
However, the improvements from bias decoding is somewhat small in
some models. The numerical gradient for training is noisy, and sometimes the SGD process might take several epochs to find
an improvement on the development set.
This happens especially on the English dataset because the base model trained with sequence level log-likelihood fits very well on the training set.
Even with the boosting trick presented during the Experiments
section, the training is still very slow. At first, we expected that the biases might give us some ideas about the trade-off between
precision and recall similar to the thresholding technique for binary classification, i.e. the based log-likelihood model might 
favors precision or recall.
However, from the analysis of the biases, we found no obvious trends favoring precision or recall.

Interestingly, the Language Model embeddings seem to have opposite effects on Spanish and English. While it is very helpful in 
English, it only degrades the performance for Spanish. The English LMs also improve convergence rate, while it is the opposite for Spanish.
We attribute this difference in the quality of the Language Model involved.
For English, the LMs are arguably better, with much lower perplexities than the LMs for Spanish. The Spanish models also have less data
to train with, and it might affect the performance.

\section{Conclusions and Future Work}
We have explored two innovations over the baseline CRF-RNN model for sequence classification: the Stacked Residual Connection, and  bias decoding. With these two improvements, it is possible to achieve state-of-the-art performance in Named Entity Recognition for both English and Spanish.

As future work, we will further investigate trainable bias decoding, and try to solve the problems presented. As the methods presented are general and language/domain independent, we plan to apply it to other domains such as health-care and expand the applications beyond NER.
%\amcomm{can we claim this is applicable to non-NER?}
%Quan: Yes, our architecture can be applied to other squence classification and chunking tasks.

\bibliography{ijcnlp2017}
\bibliographystyle{ijcnlp2017}

\end{document}